\documentclass[sigconf,nonacm=true]{acmart}
\AtBeginDocument{%
  \providecommand\BibTeX{{%
    \normalfont B\kern-0.5em{\scshape i\kern-0.25em b}\kern-0.8em\TeX}}}

\setcopyright{acmcopyright}
\copyrightyear{2018}
\acmYear{2018}
\acmDOI{10.1145/1122445.1122456}

\acmConference[Woodstock '18]{Woodstock '18: ACM Symposium on Neural
  Gaze Detection}{June 03--05, 2018}{Woodstock, NY}
\acmBooktitle{Woodstock '18: ACM Symposium on Neural Gaze Detection,
  June 03--05, 2018, Woodstock, NY}
\acmPrice{15.00}
\acmISBN{978-1-4503-XXXX-X/18/06}

\copyrightyear{2025}
\acmYear{2025}
\setcopyright{acmlicensed}\acmConference[GECCO '25]{Genetic and Evolutionary Computation Conference}{July 14--18, 2025}{Malaga, Spain}
\acmBooktitle{Genetic and Evolutionary Computation Conference (GECCO '25), July 14--18, 2025, Malaga, Spain}
\acmDOI{10.1145/3712256.3726336}
\acmISBN{979-8-4007-1465-8/2025/07}

\usepackage{enumitem}
\usepackage{ulem}

\begin{document}

\title{The Pitfalls of Benchmarking in Algorithm Selection: What We Are Getting Wrong}



\author{Gašper Petelin}
\affiliation{
  \institution{Computer Systems Department\\ Jo\v{z}ef Stefan Institute\\ Jo\v{z}ef Stefan International Postgraduate School}
  \city{Ljubljana} 
  \country{Slovenia}
}
\email{gasper.petelin@ijs.si}

\author{Gjorgjina Cenikj}
\affiliation{
  \institution{Computer Systems Department\\ Jo\v{z}ef Stefan Institute\\ Jo\v{z}ef Stefan International Postgraduate School}
  \city{Ljubljana} 
  \country{Slovenia}
}
\email{gjorgjina.cenikj@ijs.si}

\renewcommand{\shortauthors}{Petelin and Cenikj}

\begin{abstract}
Algorithm selection, aiming to identify the best algorithm for a given problem, plays a pivotal role in continuous black-box optimization. A common approach involves representing optimization functions using a set of features, which are then used to train a machine learning meta-model for selecting suitable algorithms. Various approaches have demonstrated the effectiveness of these algorithm selection meta-models. However, not all evaluation approaches are equally valid for assessing the performance of meta-models. We highlight methodological issues that frequently occur in the community and should be addressed when evaluating algorithm selection approaches. First, we identify flaws with the "leave-instance-out" evaluation technique. We show that non-informative features and meta-models can achieve high accuracy, which should not be the case with a well-designed evaluation framework. Second, we demonstrate that measuring the performance of optimization algorithms with metrics sensitive to the scale of the objective function requires careful consideration of how this impacts the construction of the meta-model, its predictions, and the model's error. Such metrics can falsely present overly optimistic performance assessments of the meta-models. This paper emphasizes the importance of careful evaluation, as loosely defined methodologies can mislead researchers, divert efforts, and introduce noise into the field.
\end{abstract}




\maketitle

\section{Introduction}

Benchmarking plays a crucial role in the development of algorithms and techniques across various fields, providing a standardized framework for performance evaluation. However, poorly designed evaluation practices can lead to misleading results~\cite{hooker1995testing, wu2021current, hewamalage2023forecast, northcutt2021pervasive, beiranvand2017best}. In this paper, we focus on algorithm selection (AS), a central topic in continuous black-box optimization~\cite{kerschke2019automated}. The goal of AS is to determine the most suitable optimization algorithm for a given problem. The most common approach to addressing this challenge begins by representing optimization problems from a benchmark suite using features, such as Exploratory Landscape Analysis (ELA)~\cite{mersmann2011exploratory} or one of the other existing feature sets~\cite{cenikj2024cross}. Next, optimization algorithms from the algorithm portfolio are evaluated on each problem individually to obtain their performance. Subsequently, a machine learning meta-model is trained to predict the most suitable optimization algorithm for a new problem based on its feature representation.

At the time of writing, the practices for performing and evaluating AS are not entirely standardized within the community. As a result, researchers employ various strategies, as there is no single, definitive way to evaluate AS methodologies~\cite{tanabe2022benchmarking}. Practitioners are free to choose from benchmark suites such as COmparing Continuous Optimizers (COCO)~\cite{hansen2021coco} or IEEE Congress on Evolutionary Computation (CEC) Special Sessions and Competitions on Real-Parameter Single-Objective Optimization~\cite{cec2017}, and often select algorithm portfolios based on the availability of code for each specific optimization algorithm. For the meta-model, standard tabular data models, such as random forests or neural networks, are commonly used. Additionally, practitioners can reformulate the problem as a classification task, aiming to predict the single best algorithm~\cite{tanabe2022benchmarking}, or as a multi-class classification task~\cite{vskvorc2022transfer}, where the goal is to recommend a set of algorithms. One can even treat AS as a regression task, where the objective is to predict the ranking of algorithms~\cite{petelin2024generalization}, its performance relative to the other algorithms~\cite{cenikj2024cross} or by how much the optimization algorithms will fall short of optimal values commonly known as target precision~\cite{jankovic2020landscape}.

With such a flexible methodology, there is potential for misuse, whether intentional or accidental. Past approaches have reported varying levels of success~\cite{lacroix2019limitations, jankovic2020landscape, cenikj2024cross, petelin2024generalization, andova2024predicting, kerschke2019automated, munoz2021sampling, cenikj2025landscape}. It is possible to combine the aforementioned components and develop evaluation methodologies that lack rigor, allowing nonsensical meta-models to significantly outperform baselines and existing models, purely due to flaws in the evaluation approach. In this paper, we highlight two such issues with the methodology commonly used in the community, where incorrect evaluation can lead to wrong or conflicting conclusions.

The first issue we investigate is the use of the COCO benchmark in AS. Originally designed for benchmarking optimization algorithms, COCO consists of 24 problem classes, each with multiple problem instances. These instances are often shifted or translated versions of the original problem to prevent optimization algorithms from overfitting to a specific objective function or search region. Consequently, problem instances within the same problem class most often have properties that are more similar to each other than to instances from different classes~\cite{long2023bbob, vskvorc2020understanding, renau2021towards}, resulting in a somewhat comparable performance of optimization algorithms within instances of the same class~\cite{long2023bbob}. Later adopted by the AS community, COCO has become a common benchmark for evaluating the capabilities of AS meta-models. In the context of AS on the COCO benchmark, two main evaluation techniques are used: the "leave-instance-out" (LIO) method and the "leave-problem-out" (LPO) method. With the LIO method, the train and test sets contain all 24 problem classes, with instances of each problem class being split between the train and test sets. Due to instances of the same class appearing in both the train and test sets, this approach is relatively forgiving. A more challenging evaluation is the LPO evaluation method, where the train set typically contains instances from 23 problem classes, and the test set contains instances from the remaining problem class. In our paper, we demonstrate that the LIO methodology, commonly used with COCO, is flawed when used to benchmark AS meta-models, as it can lead to misleading results. Meta-models achieve high performance due to spurious correlations between features and the target, rather than genuine predictive capability.

The second issue, which we believe is common, is the use of scale-sensitive metrics to measure the performance of optimization algorithms and the subsequent use of these metrics as targets to be predicted by the meta-models. This approach can lead to several issues. For instance, when problem instances are not on the same scale, as is the case with the COCO benchmark, comparing and aggregating these metrics without accounting for scale becomes problematic. Furthermore, scale-sensitive metrics can mislead meta-model construction, as the models may incorrectly attribute importance to features influenced by scale. Finally, developing baselines for meta-models trained on such metrics is challenging and can result in misleading outcomes, where meta-models appear to outperform baselines significantly due to the properties of the metrics rather than genuine model improvements.

\textbf{Our contribution:} In this paper, we identify two significant shortcomings commonly observed in the evaluation of landscape features and meta-models. Specifically, we highlight issues related to the evaluation of LIO and the use of scale-sensitive metrics for meta-model construction. For both methodological shortcomings, we provide comprehensive explanations outlining the severe weaknesses of the current methodologies and propose potential strategies for their improvement. We argue that the existing benchmarking practices for landscape features and meta-models are flawed, leading to potentially erroneous conclusions. Due to these shortcomings, we believe that some evaluations of landscape features and algorithm selection meta-models are likely unreliable due to flawed evaluation techniques.

\textbf{Outline:} The structure of the following sections is organized as follows: Section~\ref{sec:lio} examines the application of the LIO methodology within the context of AS meta-model evaluation. This is followed by Section~\ref{sec:scale_metrics}, which addresses challenges associated with using scale-sensitive metrics in model evaluation. Finally, Section~\ref{sec:conclusion} presents concluding observations and remarks.

\textbf{Reproducibility:} The experiments conducted can be replicated using the code available in the Github repository, accessible at~\cite{benchmarking_pitfalls_code}.

\section{Leave-Instance-Out Evaluation}
\label{sec:lio}

In this section, we make the following claim: \textbf{The "leave-instance-out" evaluation is flawed, as - due to spurious correlations - it can result in non-informative features and meta-models still achieving high performance, producing misleading and over-optimistic results.} To demonstrate this, we adopt an approach where we deliberately construct non-informative features that would be deemed useless by most practitioners for the AS task. Despite this, these features are used to train and evaluate meta-models within the LIO methodology, which achieves relatively good results and significantly outperforms the single best solver. This outcome suggests either that the proposed features are, in fact, useful for AS, or that the LIO methodology itself is flawed and unsuitable for evaluating meta-models.

Before presenting the full methodology, we will illustrate the problem of spurious correlations and correlated subgroups using an example from a different domain. This example will demonstrate how these issues can significantly impact a classifier's perceived accuracy. A well-known example from the computer vision domain highlights the challenges of spurious correlations when building machine learning models. In the "Husky vs. Wolf" dataset, the goal is to classify whether an image contains a wolf or a husky. Initial attempts achieved high accuracy, often surpassing human experts. However, these models struggled to generalize to new images and performed poorly in real-world scenarios, failing to differentiate between wolves and huskies effectively. The reason became evident with the application of explainability techniques: most images of wolves were taken in snowy settings, while husky images rarely featured snow~\cite{ribeiro2016should}. Consequently, the models learned to identify the presence of snow rather than distinguishing the animals themselves. This example illustrates how spurious correlations, such as background features, can undermine model construction and render any accuracy improvements meaningless. It is important to note that this issue is not unique to convolutional neural networks that were used in the study. Even manually crafted features, such as a ratio of white to non-white pixels, would boost accuracy based on this irrelevant correlation, rather than capturing a meaningful distinction between the breeds.

To demonstrate the issue with the LIO methodology, consider a scenario where a new landscape feature is introduced, and we want to assess whether it is useful for AS, without being interested in other tasks such as classification. Additionally, let us make a mild assumption that the performance of the optimization algorithm differs between problem classes but is to some degree similar for the instances of the same problem class~\cite{benchmarking_pitfalls_code}. With LIO, there are four possible cases: (a) the feature is not beneficial for either AS or problem classification, (b) the feature performs well for AS but not for classification, (c) the feature performs well only for classification but not for AS, and (d) the feature performs well for both tasks. The problematic cases in LIO are (c) and (d), where the feature performs well for classification. In such cases, if problem instances are highly similar, the feature may provide a good estimate of an algorithm's performance on a specific instance, simply through memorization (i.e. based on samples identifying to what problem class instance belongs). Additionally, features that encode the problem class of a particular instance are likely to be valuable, as they enable accurate performance estimation by indicating the problem class to which the instance belongs.

\subsection{Methodology}

In this section, we outline the methodology used to evaluate the claim that spurious correlations, together with LIO, may cause over-optimistic results.

Our methodology begins by splitting problems into training and testing sets based on LIO or LPO approaches. Samples are then obtained from each problem instance to compute features, using three different feature sets: one based on ELA and two additional sets introduced below. Optimization algorithms are evaluated on the training set to assess their performance, which is captured by ranking the algorithms in terms of the objective function value of their best-found solution. Given a set of features and algorithm ranks, meta-models are subsequently trained to map the features to their respective ranks. These meta-models are then evaluated on the test set based on their ability to rank algorithms, as measured by Pairwise Ranking Error, detailed below. This methodology aligns with approaches commonly used in other studies and is similar to standard evaluations in AS. All the details about the exact methodology used are described below.

\textbf{Benchmark suite} used in this study is the COCO benchmark, as introduced earlier. We employ the COCO benchmark suite to highlight potential pitfalls in its application for evaluating AS techniques. Specifically, we use all 24 problem classes, selecting the first 15 problem instances from each class. In this study, we only focus on $5D$ problems. Note also that other benchmarks do not use the concepts of problem classes and instances, so they may not be applicable.

\textbf{Optimization algorithm portfolio} used in this paper is defined as $\mathcal{A} = \{a_1, \dots, a_k\}$ and includes the following algorithms from the \textit{pymoo}~\cite{pymoo} framework: Genetic Algorithm (GA)~\cite{katoch2021review}, Differential Evolution (DE)~\cite{price2006differential}, Particle Swarm Optimization (PSO)~\cite{kennedy1995particle}, Evolutionary Strategy (ES)~\cite{back2005evolution}, and Covariance Matrix Adaptation Evolution Strategy (CMA-ES)~\cite{hansen2001completely}. All the algorithm hyperparameters were configured using the default values defined in the \textit{pymoo}~\cite{pymoo} framework version \textit{0.5.0}.

\textbf{Optimization algorithm performance} is measured by running each optimization algorithm 30 times on each problem instance to obtain its mean rank, which is subsequently used as the target of a meta-model. All algorithms are evaluated using 1000 function evaluations per dimension (i.e., fixed-budget). For our 5-dimensional problems, this results in a total of 5000 function evaluations.

\textbf{Feature sets} utilized in this paper are derived using three distinct approaches. Candidate solutions are generated through LHS with a sample size of $250D$, where fitness values are scaled between 0 and 1 prior to feature extraction. The three techniques employed for feature extraction are: 

\textit{i)} The first is the well-known ELA feature set~\cite{prager2023pflacco}, with a total of 85 features used to construct a meta-model. 

\textit{ii)} The second uses the so-called non-informative features. These features are designed to identify the class of a problem but are not useful for performing AS. We design non-informative features in the following way. First, $n$ candidate solutions $\mathbb{X} = \{\vec{x}_1, \dots, \vec{x}_n\}$ are obtained and evaluated $\mathbb{Y} = \{f(\vec{x}_1), \dots, f(\vec{x}_n)\}$, same as with ELA features. In the next step, the vector of fitness values $Y$ is transformed into a feature $f_i$ using the following feature construction template $f_i=agg(tr(sc\mathbb{Y}))$. In this case, $sc$ is a randomly selected scalar, $tr$ is a randomly selected transformation, and $agg$ is an aggregate function. This is repeated to obtain multiple features by repeatedly creating functions by randomly selecting aggregate function $agg \in \{mean(x), $ $median(x), $ $std(x), $ $quantile_{5}(x), $ $quantile_{25}(x), $ $quantile_{75}(x), $ $quantile_{95}(x)\}$, transformation function $tr \in \{sin(x), cos(x), x^\frac{1}{6}, x^\frac{1}{3}, x^\frac{1}{2}, x^2, log(x+1)\}$, and scalar $sc \in \{0.2, 0.3, 0.5, 0.7, 1, 2, 3, 5, 7, 9\}$. These features do not use any information about where candidate solutions were sampled $\mathbb{X}$ but only the fitness values of candidate solutions $\mathbb{Y}$. Additionally, the first feature uses the same randomly selected scalar, aggregate, and transformation functions across all objective functions, and the same applies to the second feature, etc. Although this methodology allows us to construct as many features as we like, we created the same number of features as were used with ELA.

\textit{iii)} Lastly, to demonstrate the worst-case scenario, we add an additional feature set. This feature set contains only one feature: the problem class to which each individual instance belongs. This serves mainly as a demonstration of what happens if features are strongly (or perfectly) correlated with the problem class and how a non-informative feature, in the context of AS, can achieve extremely good results with LIO evaluation.

\textbf{Meta-models} in our study minimize the MSE between the ground truth and the predicted ranks. They utilize the random forest regressor from the \textit{scikit-learn} library~\cite{scikit-learn} with default hyperparameters. Meta-models used in this section are the following: 

\textit{i)} The \texttt{random} meta-model, which randomly orders optimization algorithms; 

\textit{ii)} The \texttt{mean} meta-model, which predicts the mean rank of each optimization algorithm, functioning similarly to a single best solver in the context of ranking; 

\textit{iii)} The \texttt{ela} meta-model, where ELA features are used to predict ranks; 

\textit{iv)} The \texttt{non-inf} meta-model, where non-informative features capture basic and relatively non-informative statistics about objective function values and use them to predict ranks; and 

\textit{v)} The \texttt{class} meta-model, which uses the class of a problem instance as a feature in rank prediction.

\textbf{Meta-models' error metrics} used to evaluate the accuracy of a meta-model is the Pairwise Ranking Error (PRE) defined as:

\begin{equation}
    \label{eq:error}
    PRE = \frac{1}{|\mathcal{A}| \cdot (|\mathcal{A}| - 1)} \sum_{(a_i, a_j) \in \mathcal{A}^2, \, a_i \neq a_j} r(a_i, a_j)
\end{equation}

\begin{equation}
    r(a_i, a_j) = \begin{cases}
    0 & \text{ if } r_p(a_i, a_j) = r_g(a_i, a_j)\\ 
1 & \text{ if } r_p(a_i, a_j) \neq r_g(a_i, a_j)
\end{cases}
\end{equation}

In this context, let $a_i$ and $a_j$ represent algorithms within the portfolio, with $r_p(a_i, a_j)$ and $r_g(a_i, a_j)$ serving as functions that denote the rank differences between these optimization algorithms. Specifically, $r_p(a_i, a_j)$ outputs values of -1, 0, or 1, indicating whether the predicted rank of algorithm $a_i$ is lower than, equal to, or higher than the rank of algorithm $a_j$, respectively. Similarly, $r_g(a_i, a_j)$ follows the same rule, but instead relies on rankings based on the ground truth. This approach ensures that $r_g(a_i, a_j)$ yields values of -1, 0, or 1 by referencing the actual rankings established through 100 runs of each optimization algorithm. The PRE metric evaluates the effectiveness of ranking algorithms by quantifying the percentage of pairs of algorithms where the order of ranks between them was guessed correctly. A PRE value of 0.0 signifies a perfect match between the predicted and true order. When ranks are assigned randomly, the error value is 0.5, indicating that half of the item pairs are correctly ordered. A PRE of 1.0 represents the extreme case where all pairs are misordered, reflecting a scenario in which the predicted ranking is entirely reversed from the true order.

\subsection{Results}

\begin{figure}[!ht]
\centering
\includegraphics[width=0.48\textwidth]{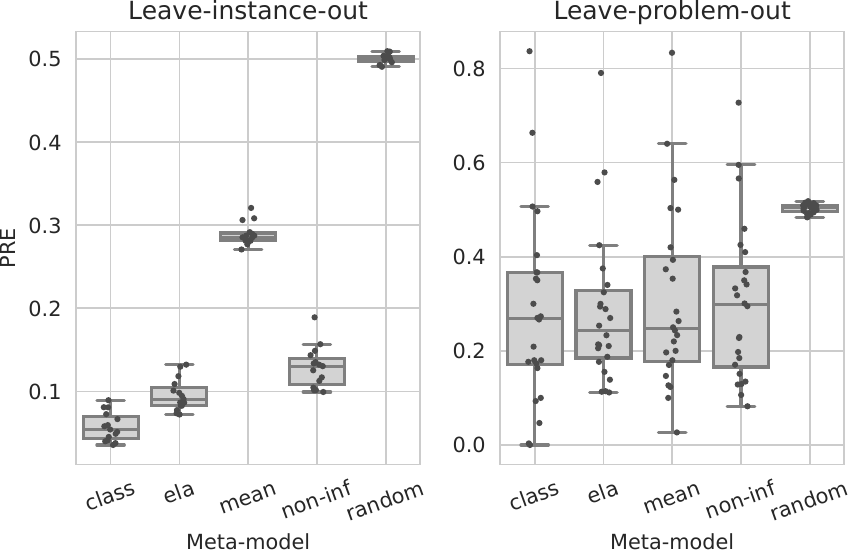}
\caption{PRE of different meta-models with the LIO evaluation strategy (top) and LPO evaluation strategy (bottom). Each dot represents one train/test split of the data.}
\label{fig:rank_errors}
\end{figure}

This subsection presents the results of evaluating various meta-models and feature sets for both the LIO and LPO methodologies. Figure~\ref{fig:rank_errors} show the PRE of different meta-models across two evaluation methodologies. We begin with the LIO evaluation. As expected, the \texttt{random} meta-model performs the worst, with a PRE of approximately 0.5, followed by the \texttt{mean} meta-model, which outputs the mean ranks of optimization algorithms calculated on the training set, achieving a PRE of around 0.3. The remaining meta-models—\texttt{ela}, \texttt{non-inf}, and \texttt{class}—substantially outperform the baselines, each achieving a PRE below 0.15. Among these three meta-models, the \texttt{non-inf} meta-model has the highest PRE at approximately 0.13, followed by the ELA-based meta-model at 0.1. However, the best-performing meta-model is \texttt{class}, which uses a single feature—the problem instance class—and achieves a PRE of 0.05. This suggests that even trivial summary statistics in the \texttt{non-inf} features can yield performance close to that of meta-models using ELA features. Furthermore, theoretically, if a feature could perfectly identify the problem class of an instance, such a meta-model would outperform the ELA meta-model. This strongly indicates that within the LIO framework, features effective in classification can significantly reduce meta-model error, even if the meta-model has no practical application for AS, as is the case with the problem class feature. For completeness, LPO validation is also presented. With LPO, all meta-models, except the \texttt{random}, achieve comparable performance, revealing that while class and non-informative features may show low PRE or even outperform ELA under LIO, this is due to LIO's tendency to reward irrelevant features that correlate with the target—a phenomenon not observed in the LPO evaluation.

\subsection{Discussion}

Here, we discuss the results from the previous subsection. As expected, we observe entirely contradictory outcomes between LIO and LPO, as one evaluation technique is much more challenging than the other~\cite{tanabe2022benchmarking}. However, what is unexpected is how well the meta-model using non-informative and class features performs with LIO. We believe the main reason for this is the presence of spurious features—features that reflect some property of a problem which, while not directly helpful for AS, may correlate with performance due to similarities between the problems in the training and test sets. Instances belonging to the same problem class can have extremely similar features, and in most cases, optimization algorithms perform similarly on instances of the same class~\cite{benchmarking_pitfalls_code}. With that in mind, there is a high likelihood that spurious correlations substantially improve the performance of meta-models without any actual progress. Any meta-model that achieves accuracy between that of \texttt{mean} and \texttt{class} cannot be claimed to perform AS due to identifying important problem properties, but rather due to spurious correlations. Although we do not present the results here due to space constraints, training meta-models using a single feature for both ELA and the proposed features, and then evaluating them on LIO, reveals a strong correlation between classification accuracy and the PRE of the meta-model. This suggests a potential similarity between classification and AS in the context of LIO tasks. In other words, it indicates that features deemed important for classification might erroneously be considered important for AS when LIO is used as the evaluation metric.

To draw a parallel with the computer vision example from the beginning of the section: features that are able to perform AS simply because they correlate with the class would be conceptually similar to the ratio of white pixels in the "Husky vs. Wolf" example. With such a feature, differentiating between wolves and huskies is not possible. However, given this feature alongside a flawed evaluation methodology, achieving high performance becomes much easier. In essence, non-informative features can significantly boost performance when faulty evaluation techniques are used. 

We would also like to emphasize that the methodology presented here cannot, in general, be used to test whether feature sets are informative or not. If a feature set has similar performance characteristics to our non-informative features, that does not necessarily mean those features are also non-informative. We only show that non-informative features can perform well with the LIO methodology.

We aim to caution researchers against using LIO evaluation and accepting it as a standard, as it will likely:

$\bullet$~Produce overly optimistic results due to the meta-model capturing spurious correlations that may artificially improve performance.

$\bullet$~Allow for the publication of new, non-informative features that appear effective solely due to faulty LIO evaluation techniques, but will never generalize to objective functions that differ from those in the test set.

$\bullet$~Yield misleading conclusions on feature importance that rely solely on LIO evaluation, as these will not accurately indicate which features are crucial for generalization but will reward features that are good at identifying the class of a problem instance. Using explainability techniques and claiming that features identified as important are actually relevant for algorithm selection in general is misleading. For example, blindly applying feature explainability techniques on a dataset where the goal is to differentiate between wolves and huskies and overgeneralizing could lead to the following conclusion: what differentiates huskies and wolves is not domestication, appearance, or behavior, but the snowy background.

The argument that LIO serves as a simpler stepping stone towards LPO is misguided. While achieving strong performance on LIO may seem like progress, it does not inherently translate to improved LPO outcomes. COCO, with its diverse yet limited set of problems, is not ideally suited as a benchmark for AS in the context of LPO due to insufficient problem quantity and dissimilarity among them. Consequently, treating LIO as an easier benchmark alternative to LPO can yield misleading results, as demonstrated in this section. If LIO is used, the authors should clearly demonstrate why their problem set is unique and justifies the use of LIO methodology, especially in cases where the training and test set instances represent the same objective function but may be slightly shifted, scaled, or rotated. While we do not address other tasks, we believe the same is true for high-level property prediction where exactly the same problems with spurious correlations can arise.

\section{Scale-Sensitive Metrics}
\label{sec:scale_metrics}

This section focuses on the metrics used to measure the performance of optimization algorithms and how they affect meta-models. We make the following claim: \textbf{Scale-sensitive metrics used to evaluate the performance of optimization algorithms, in combination with certain meta-models, can produce highly misleading results. Using objective functions with different scales may show a substantial improvement in the meta-model's performance, even when this is not the case.}

We will first define a more concrete definition of what we mean by scale-sensitive and scale-independent metrics, as to the best of our knowledge, this has not been defined in the context of AS. We define a scale-sensitive performance metric to be one that changes when the objective function is rescaled. In contrast, a scale-independent metric for comparing algorithm performance remains unchanged even when the scale of the problem changes. For example, target precision~\cite{jankovic2020landscape}, computed as the difference between best-found values and the optimal value of the objective function (defined below) is strongly correlated with the scale of the problem, as rescaling the problem will almost certainly alter the difference between the optimum value and the best-found solution. On the other hand, assuming the trajectory of optimization algorithms is not affected by the rescaling of the objective function, a metric such as an algorithm's rank would be much more stable if the problem is rescaled. However, be aware that precise definitions can be tricky. While most optimization problems today, including those discussed in the paper, are relatively invariant to the problem's scale (i.e., differently scaled objective functions do not significantly alter the algorithm's search trajectory), this is not always the case~\cite{long2023bbob}. A metric might be sensitive to scale with one portfolio of algorithms but not with another.

A similar concept of metric sensitivity to scaling has been thoroughly explored in other domains such as time series forecasting~\cite{strom2024performance}. For example, let us consider the case of a bank with two individuals forecasting the price of oil. One is based in Europe and makes forecasts in euros, while the other is based in Japan and makes forecasts in yen. Both individuals use an identical forecasting model to predict the future price of oil. Their future predictions are evaluated based on the mean absolute error (MAE). Since 100 yen is approximately 0.6€ (assuming this exchange rate remains fixed), the same forecasting model will yield different MAE values depending on the currency used for evaluation. A 1\% forecasting error might translate to a difference of a few hundred yen, but when measured in euros, it would correspond to only a few euros. Such an evaluation technique presents several challenges. Since both time series operate on completely different scales, their errors will likely reflect this disparity. For instance, an MAE of 20 might be negligible for forecasts in yen but substantial for forecasts in euros. Additionally, simply aggregating the errors without accounting for the scale of the time series would produce meaningless results. Using scale-sensitive metrics like MAE makes it impractical to compare errors across time series with different scales. How should a bank aim to maximize profit, allocate further funds, and measure the success of multiple individuals forecasting different securities? If MAE is used without accounting for differences in exchange rates, all funds would likely be allocated to the European individual simply because their MAE appears lower. Such a performance metric is clearly flawed in this context. When hundreds of securities and currencies are involved, the bank must develop a strategy to incorporate this knowledge into its decision-making process effectively. The same principle applies to AS meta-models combined with scale-sensitive metrics, where similar issues could arise due to the inappropriate handling of scale differences.

When a meta-model is trained to predict a scale-sensitive performance metric, such as the MAE, using a time series represented by a set of features, its task becomes quite tricky. A high predicted MAE could either indicate that the time series is hard to forecast or simply that the series has a larger scale. In this case, the scale-sensitive nature of the performance metric means its value can vary significantly, not only based on forecasting difficulty but also on the scale of the series itself. Thus, a time series that is harder to forecast might still have a smaller MAE than a series that is trivially forecastable but has a larger scale. This makes the meta-model's job more complex, as it must learn to separate the effects of scale from the inherent forecasting difficulty.

Additionally, a consideration with scale-sensitive metrics and time series of different scales is the need to carefully construct the baseline for a meta-model. But what would be a reasonable baseline meta-model in our case of predicting the performance of forecasting algorithms? The simplest baseline we could construct is a meta-model that predicts the mean MAE observed during the training phase. But is such a baseline really reasonable? If MAE accounts for both the scale of the time series and its difficulty in being forecasted, then such a baseline is rather meaningless. Any meta-model that can learn how a time series is scaled will outperform the scale-agnostic baseline that only predicts the mean MAE independent of the problem scale. For that reason, scale-sensitive metrics for measuring the performance of forecasting algorithms are almost never used for meta-model construction~\cite{petelin2023towards}.

\subsection{Methodology}

To demonstrate how the use of scale-sensitive metrics for evaluating optimization algorithm performance can influence the construction of meta-models and highlight how simply interpreting meta-model errors can be misleading, the following methodology is employed: a meta-model is trained to predict an optimization algorithm's target precision using a single feature that is correlated with the scale of the objective function. For the LPO evaluation, this meta-model is compared against a baseline model that predicts the mean target precision to determine whether it outperforms the baseline. Additionally, alongside meta-models predicting target precision, another set of meta-models is trained to predict the ranks of optimization algorithms. All the meta-models are then compared to evaluate if there are any discrepancies between the use of scale-sensitive and scale-free metrics. Any substantial differences between the metrics would indicate that, in one case, meta-models might achieve smaller errors simply due to learning the strong correlation between scale and target precision through the use of scale-sensitive features. 

The methodology is largely the same as in the previous section, with the \textbf{Benchmark suite} and \textbf{Optimization algorithm portfolio} remaining the same. Note that at no stage of our methodology did we intentionally rescale the COCO benchmark problems. We keep them as they are in the COCO framework with large differences in scale between problem classes. The major difference with this approach lies in the way we measure performance and the features used to train the meta-models.

\textbf{Optimization algorithm performance} is measured using a metric called target precision, which evaluates how close the optimization algorithm gets to the optimal value. In other words, optimization algorithms are assessed in terms of target precision, defined as $f(\vec{x}_{best\_found}) - f(\vec{x}_{opt})$, where $f(\vec{x}_{best\_found})$ represents the best value found by a specific algorithm, and $f(\vec{x}_{opt})$ denotes the optimum value for the given problem. This value is then used as a target for meta-model training. All the other metrics are the same as in the previous subsection.

\textbf{Feature set} in this section consists of only a single feature named $f_{scale}$, constructed in the following way. Given candidate solutions that are evaluated to obtain $Y = \{f(\vec{x}_1), \dots, f(\vec{x}_n)\}$, the feature $f_{scale}$ is defined as $f_{scale} = \mathit{max(\mathbb{Y})} - \mathit{min(\mathbb{Y})}$. This feature essentially captures how the problem is scaled based on the initial set of samples.

\sloppy
\textbf{Meta-models} used in this section are as follows: \texttt{mean-precision}, which predicts the mean precision obtained on the training data without using any features; \texttt{mean-rank}, which is identical to the previous model but predicts ranks instead of precision; and \texttt{rf-precision} and \texttt{rf-rank}, which predict precision and rank, respectively, based on a single feature defined above. All meta-models are set up to minimize MSE between predicted target precision and ground truth target precision, or, in the case of ranks, between ground truth ranks and predicted ranks.

\sloppy
\textbf{Meta-models' error metric} is defined as follows. Meta-models trained to predict target precision (\texttt{rf-precision} and \texttt{mean-precision}) are evaluated using MSE on raw predictions and PRE when target precision values are transformed into ranks. On the other hand, rank prediction meta-models (\texttt{rf-rank} and \texttt{mean-rank}) are evaluated solely using PRE.

\subsection{Results}

\begin{figure}[!ht]
\centering
\includegraphics[width=0.45\textwidth]{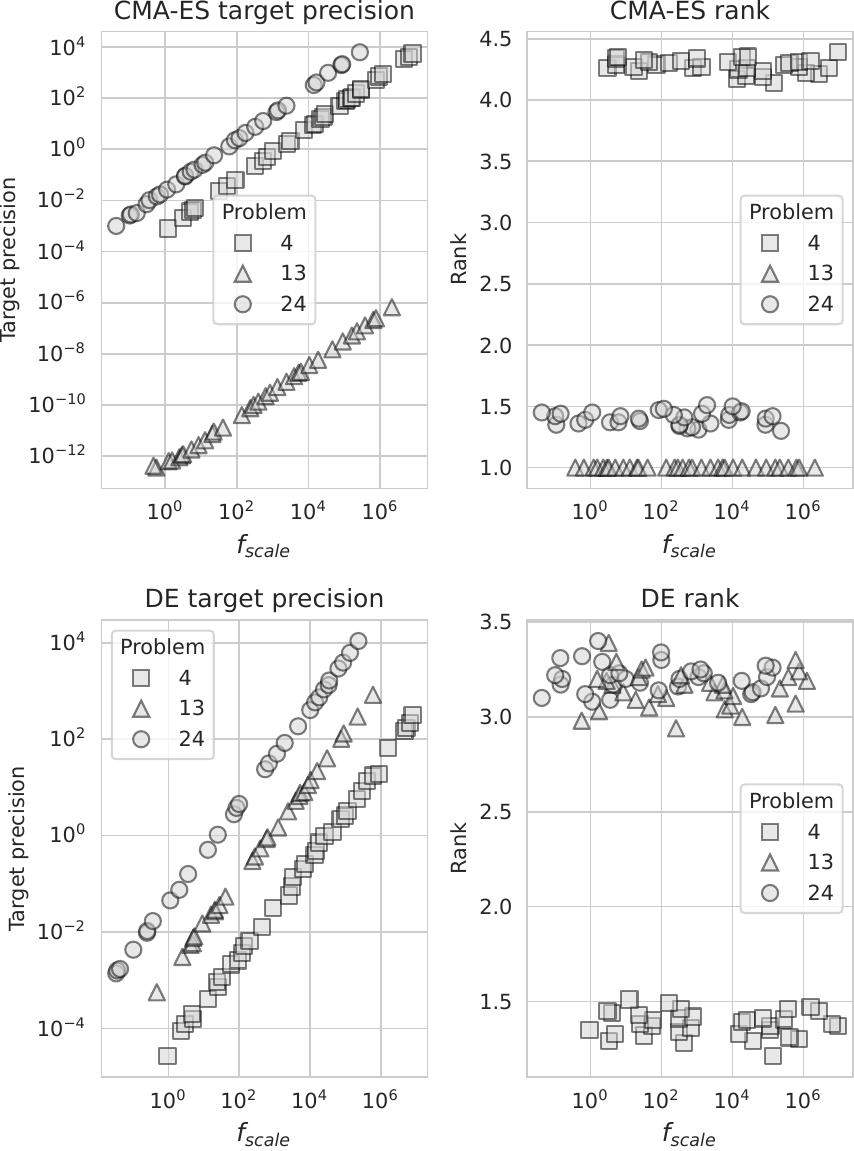}
\caption{Comparison of target precision and rank metrics (y-axis) for quantifying the performance of algorithms for the first instance of COCO problems 4, 13, and 24. All instances are artificially rescaled and their scale is measured with $f_{scale}$ feature (x-axis). Results are reported for only two out of five algorithms in the portfolio.}
\label{fig:rank_vs_precision}
\end{figure}

The first part of the results section addresses the sensitivity of metrics to scaling and how this affects scale-sensitive and scale-free metrics. Figure~\ref{fig:rank_vs_precision} illustrates this characteristic of the metrics. We plot target precision and rank relative to scale for two of the five optimization algorithms on the first instances of problems classes 4, 13, and 24. Clear patterns emerge, showing that, in this scenario, the precision metric is highly correlated with the scale of the instance, while rank remains relatively consistent, regardless of scale. This points to the fact that one can arbitrarily increase/decrease the target precision singly by rescaling the objective function, while the same is not true for ranks. Additionally, using a target precision can be a poor indicator of problem difficulty as comparisons in terms of target precision between problems on different scales (as is the case with COCO problems) are meaningless.

\begin{figure}[!ht]
\centering
\includegraphics[width=0.45\textwidth]{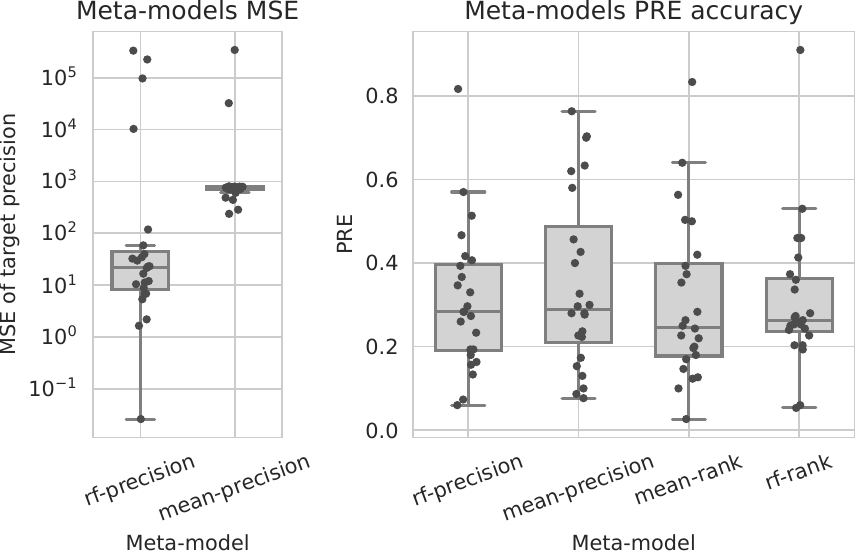}
\caption{Left: MSE calculated between true and predicted target precision for two meta-models aggregated across all five optimization algorithms. With only a single feature, the \texttt{rf-precision} meta-model outperforms the \texttt{mean-precision} by orders of magnitude. Right: Evaluating four different meta-models to determine their success in ranking five optimization algorithms in the LPO setting. Two of the meta-models directly predict the ranks, while the remaining two meta-models predict target precision, which is then transformed into ranks. The large differences between the meta-models are now nonexistent.}
\label{fig:lpo_precision_pre}
\end{figure}

In the second part of this section, we focus on the discrepancy between scale-sensitive and scale-free metrics and why setting up a baseline is so difficult with scale-sensitive metrics. First, we plot the MSE of \texttt{mean-precision} and \texttt{rf-precision} meta-models in Figure~\ref{fig:lpo_precision_pre} using the LPO methodology. The MSE is aggregated over all five optimization algorithms, and each individual point shows the error of a meta-model on one of the 24 splits. From this, we can conclude that the \texttt{rf-precision} meta-model, using only a single feature $f_{scale}$ defined above, gives extremely good results. Its accuracy is an order of magnitude better than the baseline \texttt{mean-precision}. This is confirmed with the Wilcoxon signed-rank test (p < 0.05), where we observe a statistically significant difference in the performance of the two meta-models. But how is it possible to achieve such impressive results using only a single feature, when even more advanced meta-models~\cite{tanabe2022benchmarking, cenikj2024cross, cenikj2025landscape} struggle to achieve this? And are these results really as impressive as they appear? Surely, such a good target prediction meta-model should be capable of correctly ranking optimization algorithms from best to worst, even for the LPO methodology.

The next step is to use the \texttt{rf-precision} and \texttt{mean-precision} meta-models to rank optimization algorithms. This is performed as follows: among the five algorithms, the one with the lowest predicted target precision is ranked as the best, the second lowest target precision as the second best, and so on for all five. The obtained ranks are then compared with the PRE metric to measure the error between the predicted and ground truth ranks of the optimization algorithms. Figure~\ref{fig:lpo_precision_pre} shows the PRE of all four meta-models. To our surprise, the \texttt{rf-precision} meta-model is no longer effective when evaluated with a scale-free metric. Even more surprisingly, there is no longer any difference in performance between the four meta-models. Using the Friedman test for statistical comparison at a significance level of 0.05, the performances do not differ in any significant way. So how could \texttt{rf-precision} meta-model go from significantly outperforming the baseline when evaluated with the MSE when predicting a scale-sensitive target precision metric to showing no difference between meta-models on a scale-free metric? We discuss this in the next section.

\subsection{Discussion}

Achieving a target precision of 0.03 in an optimization problem is hard to evaluate without knowing the objective function's scale. Precision depends on scaling; 0.03 might be negligible for a function ranging from 0 to 10,000 but unacceptably high for one ranging from 0 to 1. Thus, target precisions from differently scaled functions cannot be directly compared without normalization, as they represent varying relative accuracies and risk misleading conclusions.

A similar issue arises when predicting target precision errors using a meta-model. For objective functions of different scales, one can expect the errors to follow a similar pattern, meaning the meta-model will have larger errors for objective functions with larger scales. This implies that reporting metrics like MSE or MAE for a meta-model's predictions of target precision compared to the true target precision would also be meaningless without considering scale. When performing k-fold cross-validation, different train/test splits could result in errors that differ by order of magnitude, making the evaluation inconsistent and unreliable without proper scaling. 


The second issue we discuss is the contradictory results between using scale-sensitive and scale-free metrics to measure the performance of AS. Both use similar baselines (mean precision and mean rank), the same features, and a similar method of constructing the meta-model (random forest), yet the results are completely different. Why is this the case? We believe the main reason is the following: the target precision metric concurrently measures two aspects. When a particular problem is harder to solve (for example, in terms of multimodality), the optimization algorithm will, in most cases, find inferior solutions compared to an unimodal problem, thereby resulting in worse performance in terms of target precision. However, this is not the sole factor contributing to an increase in error for scale-sensitive metrics. Even if the problem's difficulty remains constant, scaling the problem can also increase or decrease the precision error. Consequently, such errors have two components that are often challenging to untangle. If a meta-model is evaluated on its ability to predict target precision, there are two ways to reduce the error. The first is to learn what makes a problem more or less difficult, and adjust the expected error that the optimization algorithm will make based on that. The other way to improve the performance of such a model is to include features that capture the scale of the problem and adjust the prediction accordingly. 

Suppose we have two objective functions for which we need to predict performance, measured by target precision, without any additional information. The best we could do in this case would be to predict the mean precision calculated from the training set. However, if information on the problem’s scale is provided, we can significantly improve precision simply by predicting the mean error conditioned on the objective function’s scale. This approach can substantially reduce the error without actually improving the ability to select the best algorithm. 
The same is not true for scale-independent metrics like the PRE metric, since the rank of the optimization algorithm does not change with scale. This points to a rather alarming fact that even if the meta-model is good at predicting target precision, this will not necessarily translate into a good AS model.

One might argue that the scale-sensitive feature used here is not relevant in practice and that no one would actually use it. While this is true, many ELA features, for instance, are also sensitive to the problem’s scale~\cite{urban_ela_not_invariant}, which would similarly introduce scale-dependent information into the meta-model. To create a proper baseline for evaluating precision, the baseline must account for the problem’s scale; otherwise, it is too weak to be useful.

We believe the following recommendations should be applied when evaluating AS meta-models:

$\bullet$~Using scale-sensitive metrics to measure algorithm performance, such as target precision, can be interpreted in multiple ways. A large error may indicate that a particular objective function is either more difficult to solve or is differently scaled. Comparing algorithms based on scale-sensitive metrics is tricky for differently scaled objective functions. Aggregating such metrics should be avoided as it can lead to misleading results.

$\bullet$~A meta-model trained to predict a scale-sensitive metric of optimization algorithm performance might struggle to learn how the algorithm's performance relates to the difficulty and scale of the problem. As a result, it may only capture how the features are related to the scale of the objective function.

$\bullet$~Trained meta-models should be evaluated using appropriate metrics. Metrics like classification accuracy, PRE, or similar alternatives are more suitable in such cases, as the meta-model only needs to learn which algorithm to select, independent of scale. Reporting the success of meta-models in predicting scale-sensitive metrics (in terms of MSE, MAE, or similar metrics) is largely meaningless, as such meta-models may not necessarily perform well for the task of algorithm selection.

$\bullet$~Using any sort of explainability techniques together with meta-models trained on scale-sensitive metrics will produce misleading feature importance values, as scale-sensitive features will have their importance artificially inflated. This indicates that the larger the differences in scale, the more important the scale-sensitive features will appear.

\section{Conclusion}
\label{sec:conclusion}

This study highlights some common but often overlooked pitfalls in algorithm selection evaluation methodologies that can lead to overly optimistic and misleading results. We presented two examples where improper evaluation practices can result in incorrect conclusions. First, we demonstrated how the use of the LIO methodology applicable when using COCO benchmark can produce meaningless meta-models with seemingly good performance. This occurs due to strong correlations between features and algorithm performances when similar problem instances appear in both the training and test sets. Second, we addressed the issue of using scale-sensitive metrics, where improvements over baseline models may seem significant but are artifacts of flaws in the metrics themselves. For both issues, we provided a detailed analysis explaining their causes and outlined strategies to mitigate them. We acknowledge that no evaluation methodology is perfect. However, we believe that current approaches to algorithm selection are inadequate, with significant effort being invested in methodologies that are ineffective.

In this work, we have highlighted key methodological challenges and addressed specific gaps in the current landscape of AS. We believe that the presented pitfalls are not simply "small" errors that are to some degree expected when writing a paper. Examples of such minor issues include incorrectly reporting some of the default hyperparameter values of an algorithm or applying inappropriate statistical tests. The pitfalls addressed in this paper, if not properly handled, can invalidate many of the core claims that authors in the community often make. This discussion is not intended as a critique of any specific papers, and especially not to question the value of widely recognized benchmarks like COCO or techniques such as ELA features, both of which remain perfectly valid and have significantly advanced the field, even though they are sometimes used within the faulty methodologies we discuss. We deliberately chose not to list specific papers where the approaches we described as problematic have been employed, focusing instead on fostering constructive discussions about methodological improvements. We acknowledge that we, the authors, have also previously employed some of the approaches we now advocate against using.

Future research should focus on refining benchmarking practices through more rigorous and accurate evaluation methodologies. While this paper primarily points out the weaknesses of current approaches used in the community, further work is needed to establish best practices for reliable assessments of AS meta-models and landscape features.

\begin{acks}
This work received financial support from the following sources: the Slovenian Research Agency through research core funding (No. P2-0098), a young researcher grant (No. PR-11263) awarded to GP, and a young researcher grant (No. PR-12393) awarded to GC, as well as the European Union’s Horizon Europe program under grant agreement No. 101077049 (CONDUCTOR). We would also like to thank the reviewers for their valuable comments.

\end{acks}

\bibliographystyle{ACM-Reference-Format}
\bibliography{sample-base}


\begin{thebibliography}{34}


\ifx \showCODEN    \undefined \def \showCODEN     #1{\unskip}     \fi
\ifx \showDOI      \undefined \def \showDOI       #1{#1}\fi
\ifx \showISBNx    \undefined \def \showISBNx     #1{\unskip}     \fi
\ifx \showISBNxiii \undefined \def \showISBNxiii  #1{\unskip}     \fi
\ifx \showISSN     \undefined \def \showISSN      #1{\unskip}     \fi
\ifx \showLCCN     \undefined \def \showLCCN      #1{\unskip}     \fi
\ifx \shownote     \undefined \def \shownote      #1{#1}          \fi
\ifx \showarticletitle \undefined \def \showarticletitle #1{#1}   \fi
\ifx \showURL      \undefined \def \showURL       {\relax}        \fi
\providecommand\bibfield[2]{#2}
\providecommand\bibinfo[2]{#2}
\providecommand\natexlab[1]{#1}
\providecommand\showeprint[2][]{arXiv:#2}

\bibitem[\protect\citeauthoryear{Andova, Cork, Vodopija, Tu{\v{s}}ar, and Filipi{\v{c}}}{Andova et~al\mbox{.}}{2024}]%
        {andova2024predicting}
\bibfield{author}{\bibinfo{person}{Andrejaana Andova}, \bibinfo{person}{Jordan~N Cork}, \bibinfo{person}{Aljo{\v{s}}a Vodopija}, \bibinfo{person}{Tea Tu{\v{s}}ar}, {and} \bibinfo{person}{Bogdan Filipi{\v{c}}}.} \bibinfo{year}{2024}\natexlab{}.
\newblock \showarticletitle{Predicting Algorithm Performance in Constrained Multiobjective Optimization: A Tough Nut to Crack}. In \bibinfo{booktitle}{\emph{International Conference on the Applications of Evolutionary Computation (Part of EvoStar)}}. Springer, \bibinfo{pages}{310--325}.
\newblock


\bibitem[\protect\citeauthoryear{B{\"a}ck}{B{\"a}ck}{2005}]%
        {back2005evolution}
\bibfield{author}{\bibinfo{person}{Thomas B{\"a}ck}.} \bibinfo{year}{2005}\natexlab{}.
\newblock \showarticletitle{Evolution strategies: An alternative evolutionary algorithm}. In \bibinfo{booktitle}{\emph{Artificial Evolution: European Conference, AE 95 Brest, France, September 4--6, 1995 Selected Papers}}. Springer, \bibinfo{pages}{1--20}.
\newblock


\bibitem[\protect\citeauthoryear{Beiranvand, Hare, and Lucet}{Beiranvand et~al\mbox{.}}{2017}]%
        {beiranvand2017best}
\bibfield{author}{\bibinfo{person}{Vahid Beiranvand}, \bibinfo{person}{Warren Hare}, {and} \bibinfo{person}{Yves Lucet}.} \bibinfo{year}{2017}\natexlab{}.
\newblock \showarticletitle{Best practices for comparing optimization algorithms}.
\newblock \bibinfo{journal}{\emph{Optimization and Engineering}}  \bibinfo{volume}{18} (\bibinfo{year}{2017}), \bibinfo{pages}{815--848}.
\newblock


\bibitem[\protect\citeauthoryear{{Blank} and {Deb}}{{Blank} and {Deb}}{2020}]%
        {pymoo}
\bibfield{author}{\bibinfo{person}{J. {Blank}} {and} \bibinfo{person}{K. {Deb}}.} \bibinfo{year}{2020}\natexlab{}.
\newblock \showarticletitle{pymoo: Multi-Objective Optimization in Python}.
\newblock \bibinfo{journal}{\emph{IEEE Access}}  \bibinfo{volume}{8} (\bibinfo{year}{2020}), \bibinfo{pages}{89497--89509}.
\newblock


\bibitem[\protect\citeauthoryear{Cenikj, Petelin, and Eftimov}{Cenikj et~al\mbox{.}}{2024}]%
        {cenikj2024cross}
\bibfield{author}{\bibinfo{person}{Gjorgjina Cenikj}, \bibinfo{person}{Ga{\v{s}}per Petelin}, {and} \bibinfo{person}{Tome Eftimov}.} \bibinfo{year}{2024}\natexlab{}.
\newblock \showarticletitle{A cross-benchmark examination of feature-based algorithm selector generalization in single-objective numerical optimization}.
\newblock \bibinfo{journal}{\emph{Swarm and Evolutionary Computation}}  \bibinfo{volume}{87} (\bibinfo{year}{2024}), \bibinfo{pages}{101534}.
\newblock


\bibitem[\protect\citeauthoryear{Cenikj, Petelin, Seiler, Cenikj, and Eftimov}{Cenikj et~al\mbox{.}}{2025}]%
        {cenikj2025landscape}
\bibfield{author}{\bibinfo{person}{Gjorgjina Cenikj}, \bibinfo{person}{Ga{\v{s}}per Petelin}, \bibinfo{person}{Moritz Seiler}, \bibinfo{person}{Nikola Cenikj}, {and} \bibinfo{person}{Tome Eftimov}.} \bibinfo{year}{2025}\natexlab{}.
\newblock \showarticletitle{Landscape Features in Single-Objective Continuous Optimization: Have We Hit a Wall in Algorithm Selection Generalization?}
\newblock \bibinfo{journal}{\emph{arXiv preprint arXiv:2501.17663}} (\bibinfo{year}{2025}).
\newblock


\bibitem[\protect\citeauthoryear{Hansen, Auger, Ros, Mersmann, Tu{\v{s}}ar, and Brockhoff}{Hansen et~al\mbox{.}}{2021}]%
        {hansen2021coco}
\bibfield{author}{\bibinfo{person}{Nikolaus Hansen}, \bibinfo{person}{Anne Auger}, \bibinfo{person}{Raymond Ros}, \bibinfo{person}{Olaf Mersmann}, \bibinfo{person}{Tea Tu{\v{s}}ar}, {and} \bibinfo{person}{Dimo Brockhoff}.} \bibinfo{year}{2021}\natexlab{}.
\newblock \showarticletitle{COCO: A platform for comparing continuous optimizers in a black-box setting}.
\newblock \bibinfo{journal}{\emph{Optimization Methods and Software}} \bibinfo{volume}{36}, \bibinfo{number}{1} (\bibinfo{year}{2021}), \bibinfo{pages}{114--144}.
\newblock


\bibitem[\protect\citeauthoryear{Hansen and Ostermeier}{Hansen and Ostermeier}{2001}]%
        {hansen2001completely}
\bibfield{author}{\bibinfo{person}{Nikolaus Hansen} {and} \bibinfo{person}{Andreas Ostermeier}.} \bibinfo{year}{2001}\natexlab{}.
\newblock \showarticletitle{Completely derandomized self-adaptation in evolution strategies}.
\newblock \bibinfo{journal}{\emph{Evolutionary computation}} \bibinfo{volume}{9}, \bibinfo{number}{2} (\bibinfo{year}{2001}), \bibinfo{pages}{159--195}.
\newblock


\bibitem[\protect\citeauthoryear{Hewamalage, Ackermann, and Bergmeir}{Hewamalage et~al\mbox{.}}{2023}]%
        {hewamalage2023forecast}
\bibfield{author}{\bibinfo{person}{Hansika Hewamalage}, \bibinfo{person}{Klaus Ackermann}, {and} \bibinfo{person}{Christoph Bergmeir}.} \bibinfo{year}{2023}\natexlab{}.
\newblock \showarticletitle{Forecast evaluation for data scientists: common pitfalls and best practices}.
\newblock \bibinfo{journal}{\emph{Data Mining and Knowledge Discovery}} \bibinfo{volume}{37}, \bibinfo{number}{2} (\bibinfo{year}{2023}), \bibinfo{pages}{788--832}.
\newblock


\bibitem[\protect\citeauthoryear{Hooker}{Hooker}{1995}]%
        {hooker1995testing}
\bibfield{author}{\bibinfo{person}{John~N Hooker}.} \bibinfo{year}{1995}\natexlab{}.
\newblock \showarticletitle{Testing heuristics: We have it all wrong}.
\newblock \bibinfo{journal}{\emph{Journal of heuristics}}  \bibinfo{volume}{1} (\bibinfo{year}{1995}), \bibinfo{pages}{33--42}.
\newblock


\bibitem[\protect\citeauthoryear{Jankovic and Doerr}{Jankovic and Doerr}{2020}]%
        {jankovic2020landscape}
\bibfield{author}{\bibinfo{person}{Anja Jankovic} {and} \bibinfo{person}{Carola Doerr}.} \bibinfo{year}{2020}\natexlab{}.
\newblock \showarticletitle{Landscape-aware fixed-budget performance regression and algorithm selection for modular CMA-ES variants}. In \bibinfo{booktitle}{\emph{Proceedings of the 2020 Genetic and Evolutionary Computation Conference}}. \bibinfo{pages}{841--849}.
\newblock


\bibitem[\protect\citeauthoryear{Katoch, Chauhan, and Kumar}{Katoch et~al\mbox{.}}{2021}]%
        {katoch2021review}
\bibfield{author}{\bibinfo{person}{Sourabh Katoch}, \bibinfo{person}{Sumit~Singh Chauhan}, {and} \bibinfo{person}{Vijay Kumar}.} \bibinfo{year}{2021}\natexlab{}.
\newblock \showarticletitle{A review on genetic algorithm: past, present, and future}.
\newblock \bibinfo{journal}{\emph{Multimedia Tools and Applications}}  \bibinfo{volume}{80} (\bibinfo{year}{2021}), \bibinfo{pages}{8091--8126}.
\newblock


\bibitem[\protect\citeauthoryear{Kennedy and Eberhart}{Kennedy and Eberhart}{1995}]%
        {kennedy1995particle}
\bibfield{author}{\bibinfo{person}{James Kennedy} {and} \bibinfo{person}{Russell Eberhart}.} \bibinfo{year}{1995}\natexlab{}.
\newblock \showarticletitle{Particle swarm optimization}. In \bibinfo{booktitle}{\emph{Proceedings of ICNN'95-international conference on neural networks}}, Vol.~\bibinfo{volume}{4}. IEEE, \bibinfo{pages}{1942--1948}.
\newblock


\bibitem[\protect\citeauthoryear{Kerschke and Trautmann}{Kerschke and Trautmann}{2019}]%
        {kerschke2019automated}
\bibfield{author}{\bibinfo{person}{Pascal Kerschke} {and} \bibinfo{person}{Heike Trautmann}.} \bibinfo{year}{2019}\natexlab{}.
\newblock \showarticletitle{Automated algorithm selection on continuous black-box problems by combining exploratory landscape analysis and machine learning}.
\newblock \bibinfo{journal}{\emph{Evolutionary computation}} \bibinfo{volume}{27}, \bibinfo{number}{1} (\bibinfo{year}{2019}), \bibinfo{pages}{99--127}.
\newblock


\bibitem[\protect\citeauthoryear{Lacroix and McCall}{Lacroix and McCall}{2019}]%
        {lacroix2019limitations}
\bibfield{author}{\bibinfo{person}{Benjamin Lacroix} {and} \bibinfo{person}{John McCall}.} \bibinfo{year}{2019}\natexlab{}.
\newblock \showarticletitle{Limitations of benchmark sets and landscape features for algorithm selection and performance prediction}. In \bibinfo{booktitle}{\emph{Proceedings of the Genetic and Evolutionary Computation Conference Companion}}. \bibinfo{pages}{261--262}.
\newblock


\bibitem[\protect\citeauthoryear{Long, Vermetten, van Stein, and Kononova}{Long et~al\mbox{.}}{2023}]%
        {long2023bbob}
\bibfield{author}{\bibinfo{person}{Fu~Xing Long}, \bibinfo{person}{Diederick Vermetten}, \bibinfo{person}{Bas van Stein}, {and} \bibinfo{person}{Anna~V Kononova}.} \bibinfo{year}{2023}\natexlab{}.
\newblock \showarticletitle{BBOB instance analysis: Landscape properties and algorithm performance across problem instances}. In \bibinfo{booktitle}{\emph{International Conference on the Applications of Evolutionary Computation (Part of EvoStar)}}. Springer, \bibinfo{pages}{380--395}.
\newblock


\bibitem[\protect\citeauthoryear{Mersmann, Bischl, Trautmann, Preuss, Weihs, and Rudolph}{Mersmann et~al\mbox{.}}{2011}]%
        {mersmann2011exploratory}
\bibfield{author}{\bibinfo{person}{Olaf Mersmann}, \bibinfo{person}{Bernd Bischl}, \bibinfo{person}{Heike Trautmann}, \bibinfo{person}{Mike Preuss}, \bibinfo{person}{Claus Weihs}, {and} \bibinfo{person}{G{\"u}nter Rudolph}.} \bibinfo{year}{2011}\natexlab{}.
\newblock \showarticletitle{Exploratory landscape analysis}. In \bibinfo{booktitle}{\emph{Proceedings of the 13th annual conference on Genetic and evolutionary computation}}. \bibinfo{pages}{829--836}.
\newblock


\bibitem[\protect\citeauthoryear{Mu{\~n}oz and Kirley}{Mu{\~n}oz and Kirley}{2021}]%
        {munoz2021sampling}
\bibfield{author}{\bibinfo{person}{Mario~Andr{\'e}s Mu{\~n}oz} {and} \bibinfo{person}{Michael Kirley}.} \bibinfo{year}{2021}\natexlab{}.
\newblock \showarticletitle{Sampling effects on algorithm selection for continuous black-box optimization}.
\newblock \bibinfo{journal}{\emph{Algorithms}} \bibinfo{volume}{14}, \bibinfo{number}{1} (\bibinfo{year}{2021}), \bibinfo{pages}{19}.
\newblock


\bibitem[\protect\citeauthoryear{Northcutt, Athalye, and Mueller}{Northcutt et~al\mbox{.}}{2021}]%
        {northcutt2021pervasive}
\bibfield{author}{\bibinfo{person}{Curtis~G Northcutt}, \bibinfo{person}{Anish Athalye}, {and} \bibinfo{person}{Jonas Mueller}.} \bibinfo{year}{2021}\natexlab{}.
\newblock \showarticletitle{Pervasive label errors in test sets destabilize machine learning benchmarks}.
\newblock \bibinfo{journal}{\emph{arXiv preprint arXiv:2103.14749}} (\bibinfo{year}{2021}).
\newblock


\bibitem[\protect\citeauthoryear{Pedregosa, Varoquaux, Gramfort, Michel, Thirion, Grisel, Blondel, Prettenhofer, Weiss, Dubourg, Vanderplas, Passos, Cournapeau, Brucher, Perrot, and Duchesnay}{Pedregosa et~al\mbox{.}}{2011}]%
        {scikit-learn}
\bibfield{author}{\bibinfo{person}{F. Pedregosa}, \bibinfo{person}{G. Varoquaux}, \bibinfo{person}{A. Gramfort}, \bibinfo{person}{V. Michel}, \bibinfo{person}{B. Thirion}, \bibinfo{person}{O. Grisel}, \bibinfo{person}{M. Blondel}, \bibinfo{person}{P. Prettenhofer}, \bibinfo{person}{R. Weiss}, \bibinfo{person}{V. Dubourg}, \bibinfo{person}{J. Vanderplas}, \bibinfo{person}{A. Passos}, \bibinfo{person}{D. Cournapeau}, \bibinfo{person}{M. Brucher}, \bibinfo{person}{M. Perrot}, {and} \bibinfo{person}{E. Duchesnay}.} \bibinfo{year}{2011}\natexlab{}.
\newblock \showarticletitle{Scikit-learn: Machine Learning in {P}ython}.
\newblock \bibinfo{journal}{\emph{Journal of Machine Learning Research}}  \bibinfo{volume}{12} (\bibinfo{year}{2011}), \bibinfo{pages}{2825--2830}.
\newblock


\bibitem[\protect\citeauthoryear{Petelin and Cenikj}{Petelin and Cenikj}{2024}]%
        {petelin2024generalization}
\bibfield{author}{\bibinfo{person}{Ga{\v{s}}per Petelin} {and} \bibinfo{person}{Gjorgjina Cenikj}.} \bibinfo{year}{2024}\natexlab{}.
\newblock \showarticletitle{On Generalization of ELA Feature Groups}. In \bibinfo{booktitle}{\emph{Proceedings of the Genetic and Evolutionary Computation Conference Companion}}. \bibinfo{pages}{419--422}.
\newblock


\bibitem[\protect\citeauthoryear{Petelin and Cenikj}{Petelin and Cenikj}{2025}]%
        {benchmarking_pitfalls_code}
\bibfield{author}{\bibinfo{person}{Ga{\v{s}}per Petelin} {and} \bibinfo{person}{Gjorgjina Cenikj}.} \bibinfo{year}{2025}\natexlab{}.
\newblock \bibinfo{title}{Benchmarking Pitfalls}.
\newblock \bibinfo{howpublished}{\url{https://github.com/gasperpetelin/benchmarking-as-problems}}.
\newblock
\newblock
\shownote{Accessed: 2025-01-20}.


\bibitem[\protect\citeauthoryear{Petelin, Cenikj, and Eftimov}{Petelin et~al\mbox{.}}{2023}]%
        {petelin2023towards}
\bibfield{author}{\bibinfo{person}{Ga{\v{s}}per Petelin}, \bibinfo{person}{Gjorgjina Cenikj}, {and} \bibinfo{person}{Tome Eftimov}.} \bibinfo{year}{2023}\natexlab{}.
\newblock \showarticletitle{Towards understanding the importance of time-series features in automated algorithm performance prediction}.
\newblock \bibinfo{journal}{\emph{Expert Systems with Applications}}  \bibinfo{volume}{213} (\bibinfo{year}{2023}), \bibinfo{pages}{119023}.
\newblock


\bibitem[\protect\citeauthoryear{Prager and Trautmann}{Prager and Trautmann}{2023}]%
        {prager2023pflacco}
\bibfield{author}{\bibinfo{person}{Raphael~Patrick Prager} {and} \bibinfo{person}{Heike Trautmann}.} \bibinfo{year}{2023}\natexlab{}.
\newblock \showarticletitle{Pflacco: Feature-based landscape analysis of continuous and constrained optimization problems in Python}.
\newblock \bibinfo{journal}{\emph{Evolutionary Computation}} (\bibinfo{year}{2023}), \bibinfo{pages}{1--25}.
\newblock


\bibitem[\protect\citeauthoryear{Price, Storn, and Lampinen}{Price et~al\mbox{.}}{2006}]%
        {price2006differential}
\bibfield{author}{\bibinfo{person}{Kenneth Price}, \bibinfo{person}{Rainer~M Storn}, {and} \bibinfo{person}{Jouni~A Lampinen}.} \bibinfo{year}{2006}\natexlab{}.
\newblock \bibinfo{booktitle}{\emph{Differential evolution: a practical approach to global optimization}}.
\newblock \bibinfo{publisher}{Springer Science \& Business Media}.
\newblock


\bibitem[\protect\citeauthoryear{Renau, Dr{\'e}o, Doerr, and Doerr}{Renau et~al\mbox{.}}{2021}]%
        {renau2021towards}
\bibfield{author}{\bibinfo{person}{Quentin Renau}, \bibinfo{person}{Johann Dr{\'e}o}, \bibinfo{person}{Carola Doerr}, {and} \bibinfo{person}{Benjamin Doerr}.} \bibinfo{year}{2021}\natexlab{}.
\newblock \showarticletitle{Towards explainable exploratory landscape analysis: extreme feature selection for classifying BBOB functions}. In \bibinfo{booktitle}{\emph{Applications of Evolutionary Computation: 24th International Conference, EvoApplications 2021, Held as Part of EvoStar 2021, Virtual Event, April 7--9, 2021, Proceedings 24}}. Springer, \bibinfo{pages}{17--33}.
\newblock


\bibitem[\protect\citeauthoryear{Ribeiro, Singh, and Guestrin}{Ribeiro et~al\mbox{.}}{2016}]%
        {ribeiro2016should}
\bibfield{author}{\bibinfo{person}{Marco~Tulio Ribeiro}, \bibinfo{person}{Sameer Singh}, {and} \bibinfo{person}{Carlos Guestrin}.} \bibinfo{year}{2016}\natexlab{}.
\newblock \showarticletitle{" Why should i trust you?" Explaining the predictions of any classifier}. In \bibinfo{booktitle}{\emph{Proceedings of the 22nd ACM SIGKDD international conference on knowledge discovery and data mining}}. \bibinfo{pages}{1135--1144}.
\newblock


\bibitem[\protect\citeauthoryear{{\v{S}}kvorc, Eftimov, and Koro{\v{s}}ec}{{\v{S}}kvorc et~al\mbox{.}}{2020}]%
        {vskvorc2020understanding}
\bibfield{author}{\bibinfo{person}{Urban {\v{S}}kvorc}, \bibinfo{person}{Tome Eftimov}, {and} \bibinfo{person}{Peter Koro{\v{s}}ec}.} \bibinfo{year}{2020}\natexlab{}.
\newblock \showarticletitle{Understanding the problem space in single-objective numerical optimization using exploratory landscape analysis}.
\newblock \bibinfo{journal}{\emph{Applied Soft Computing}}  \bibinfo{volume}{90} (\bibinfo{year}{2020}), \bibinfo{pages}{106138}.
\newblock


\bibitem[\protect\citeauthoryear{{\v{S}}kvorc, Eftimov, and Koro{\v{s}}ec}{{\v{S}}kvorc et~al\mbox{.}}{2022}]%
        {vskvorc2022transfer}
\bibfield{author}{\bibinfo{person}{Urban {\v{S}}kvorc}, \bibinfo{person}{Tome Eftimov}, {and} \bibinfo{person}{Peter Koro{\v{s}}ec}.} \bibinfo{year}{2022}\natexlab{}.
\newblock \showarticletitle{Transfer learning analysis of multi-class classification for landscape-aware algorithm selection}.
\newblock \bibinfo{journal}{\emph{mathematics}} \bibinfo{volume}{10}, \bibinfo{number}{3} (\bibinfo{year}{2022}), \bibinfo{pages}{432}.
\newblock


\bibitem[\protect\citeauthoryear{Str{\o}m and Gundersen}{Str{\o}m and Gundersen}{2024}]%
        {strom2024performance}
\bibfield{author}{\bibinfo{person}{Eivind Str{\o}m} {and} \bibinfo{person}{Odd~Erik Gundersen}.} \bibinfo{year}{2024}\natexlab{}.
\newblock \showarticletitle{Performance metrics for multi-step forecasting measuring win-loss, seasonal variance and forecast stability: an empirical study}.
\newblock \bibinfo{journal}{\emph{Applied Intelligence}} \bibinfo{volume}{54}, \bibinfo{number}{21} (\bibinfo{year}{2024}), \bibinfo{pages}{10490--10515}.
\newblock


\bibitem[\protect\citeauthoryear{Tanabe}{Tanabe}{2022}]%
        {tanabe2022benchmarking}
\bibfield{author}{\bibinfo{person}{Ryoji Tanabe}.} \bibinfo{year}{2022}\natexlab{}.
\newblock \showarticletitle{Benchmarking Feature-Based Algorithm Selection Systems for Black-Box Numerical Optimization}.
\newblock \bibinfo{journal}{\emph{IEEE Transactions on Evolutionary Computation}} \bibinfo{volume}{26}, \bibinfo{number}{6} (\bibinfo{year}{2022}), \bibinfo{pages}{1321--1335}.
\newblock


\bibitem[\protect\citeauthoryear{Wu, Mallipeddi, and Suganthan}{Wu et~al\mbox{.}}{2016}]%
        {cec2017}
\bibfield{author}{\bibinfo{person}{Guohua Wu}, \bibinfo{person}{Rammohan Mallipeddi}, {and} \bibinfo{person}{Ponnuthurai Suganthan}.} \bibinfo{year}{2016}\natexlab{}.
\newblock \showarticletitle{Problem Definitions and Evaluation Criteria for the CEC 2017 Competition and Special Session on Constrained Single Objective Real-Parameter Optimization}.
\newblock \bibinfo{journal}{\emph{Computational Intelligence Laboratory, Zhengzhou University, Zhengzhou China and Technical Report, Nanyang Technological University, Singapore.}} (\bibinfo{date}{10} \bibinfo{year}{2016}).
\newblock


\bibitem[\protect\citeauthoryear{Wu and Keogh}{Wu and Keogh}{2021}]%
        {wu2021current}
\bibfield{author}{\bibinfo{person}{Renjie Wu} {and} \bibinfo{person}{Eamonn~J Keogh}.} \bibinfo{year}{2021}\natexlab{}.
\newblock \showarticletitle{Current time series anomaly detection benchmarks are flawed and are creating the illusion of progress}.
\newblock \bibinfo{journal}{\emph{IEEE transactions on knowledge and data engineering}} \bibinfo{volume}{35}, \bibinfo{number}{3} (\bibinfo{year}{2021}), \bibinfo{pages}{2421--2429}.
\newblock


\bibitem[\protect\citeauthoryear{Škvorc, Eftimov, and Korošec}{Škvorc et~al\mbox{.}}{2020}]%
        {urban_ela_not_invariant}
\bibfield{author}{\bibinfo{person}{Urban Škvorc}, \bibinfo{person}{Tome Eftimov}, {and} \bibinfo{person}{Peter Korošec}.} \bibinfo{year}{2020}\natexlab{}.
\newblock \showarticletitle{Understanding the problem space in single-objective numerical optimization using exploratory landscape analysis}.
\newblock \bibinfo{journal}{\emph{Applied Soft Computing}}  \bibinfo{volume}{90} (\bibinfo{year}{2020}), \bibinfo{pages}{106138}.
\newblock
\showISSN{1568-4946}
\urldef\tempurl%
\url{https://doi.org/10.1016/j.asoc.2020.106138}
\showDOI{\tempurl}


\end{thebibliography}

\appendix









\end{document}